\documentclass[twoside]{article}

\usepackage[accepted]{aistats2021}
\usepackage{graphicx}

\begin{document}

%

%

\twocolumn[

\aistatstitle{MISO-wiLDCosts: Multi Information Source Optimization with Location Dependent Costs}

\aistatsauthor{  Antonio Candelieri, Francesco Archetti }

\aistatsaddress{ University of Milano-Bicocca } ]

\begin{abstract}
This paper addresses black-box optimization over multiple information sources whose both fidelity and query cost change over the search space, that is they are \emph{location dependent}. The approach uses: ($i$) an Augmented Gaussian Process, recently proposed in multi-information source optimization as a single model of the objective function over search space and sources, and ($ii$) a Gaussian Process to model the location-dependent cost of each source. The former is used into a  Confidence Bound based acquisition function to select the next source and location to query, while the latter is used to penalize the value of the acquisition depending on the expected query cost for any source-location pair.
The proposed approach is evaluated on a set of Hyperparameters Optimization tasks, consisting of two Machine Learning classifiers and three datasets of different sizes.
\end{abstract}

\section{Introduction}
\subsection{Overview}
Bayesian Optimization (BO) (Shahriari et al., 2015; Frazier, 2018; Archetti and Candelieri, 2019) is a sample efficient strategy for global optimization of black-box, expensive and multi-extremal functions.
The reference global optimization problem is:
\begin{equation}
\underset{x \in \Omega \subset \Re^d}{\arg \min f(x)}
\label{eq:globopt}
\end{equation}
where $\Omega$ is the $d$-dimensional search space.

BO consists of two key components: a probabilistic surrogate model of $f(x)$, fitted on the function evaluations performed so far, and an acquisition function providing the next location to query, while balancing between \emph{exploitation} and \emph{exploration}, depending on the predictions, and associated uncertainty, of the probabilistic surrogate model. Updating the model conditioned to the observations, and selecting the next location to query, are sequentially iterated until some termination criterion is met, usually a maximum number of function evaluations.

Thanks to its sample efficiency, compared to competing methods, BO is currently the standard technique for Automated Machine Learning (AutoML) (Hutter et al., 2019; Candelieri and Archetti, 2019) and also successfully applied in Neural Architecture Search (NAS) (Elsken et al., 2019). Other relevant applications of BO concern simulation-optimization (Sha et al., 2020) and control of complex systems (Candelieri et al., 2018). However, the basic BO algorithm does not directly incorporate and use any information about query cost, which is instead crucial in many real-life applications. An example comes just from AutoML and NAS: (Strubell et al., 2020) provides an estimation about the financial and environmental costs for optimizing the hyperparameters of deep neural networks in the domain of Natural Language Processing. The astonishing amount of energy required can generate an amount of CO$_2$ emissions around five times those generated by a car during its own longlife.

Therefore, recent research studies have been proposing innovative BO-based approaches designed for more challenging settings, where information about query cost is explicitly used.
Research can be basically split into two branches: ($i$) \emph{multi-information sources optimization} (MISO), characterized by the availability of cheap approximations (i.e., sources) of the more expensive $f(x)$, and ($ii$) \emph{cost-aware} (Bayesian) optimization, assuming a cost, to query $f(x)$, which changes over the search space.

\subsection{Multi-Information Source Optimization}
As far as MISO is concerned, the problem (\ref{eq:globopt}) has to be solved by efficiently using $S$ information sources, $f_1(x),\dots,f_S(x)$, providing approximations, with different accuracy, of $f(x)$, each one at its own query cost $c_s$, with $s=1,\dots,S$. Thus accuracy is location dependent, but cost is not. Efficiently solving (\ref{eq:globopt}) means that the query cost cumulated along the optimization process must be kept as low as possible. Sources can be always sorted by decreasing query cost, so that $f_1(x)$ is the most expensive one, usually with $f_1(x)=f(x)$ in the case that $f(x)$ can be directly queried.

When information sources come with an explicit information about their approximation quality (aka \emph{fidelity}), MISO specializes in \emph{multi-fidelity} optimization, first proposed in (Kennedy and O'Haga, 2000) and more recently addressed in (Peherstorfer et al., 2017; Sen et al., 2018; Marques et al., 2018; Chaudhuri et al., 2019; Kandasamy et al., 2019; Song et al., 2019). Most multi-fidelity approaches exploit hierarchical relations among sources, based on their fidelities, even if already in (March and Willcox., 2012), some drawbacks were highlighted. More precisely, hirerachical organization requires the assumption that information sources are unbiased, meaning that noise must be independent across sources (Lam et al., 2015; Poloczek et al., 2017). Moreover, querying a source with a certain fidelity at a given location $x$ implies that no further knowledge can be obtained by querying any other source of lower  fidelity, at any location. Consequently, most of the the multi-fidelity approaches cannot be applied in the case that sources cannot be hierarchically organized.

In (Lam et al., 2015) the first approach for non-hierarchical information sources has been proposed, addressing \emph{location-dependent} fidelities of the sources and defining the more general MISO setting. More recently, (Poloczek et al., 2017; Ghoreishi and Allaire, 2019) have been provided improvements to (Lam et al., 2015). All these methods are based on the idea of using a separate model for each information source (i.e., a Gaussian Process - GP) and then fusing their predictions and related uncertainties through the method proposed in (Winkler, 1981), which became the standard practice for the fusion of normally distributed data. In (Candelieri et al., 2020) a different procedure has been recently proposed, where instead of \textit{fusing} GPs, \textit{sparsification} is used to create the so-called Augmented Gaussian Process (AGP). We used this model -- summarized in Sec. \ref{sec:agp} -- and extended MISO-AGP to deal with location-dependent costs. 

In AutoML, MISO and multifidelity have been first considered in (Swersky et al., 2013), proposing an approach able to use small datasets to quickly optimize the hyperparameters of a Machine Learning (ML) algorithm on a large dataset. More recently,
(Klein et al., 2017) proposed FABOLAS (FAst Bayesian Optimization on LArge dataSets), a hyperparameters optimization (HPO) tool that simultaneously optimizes the hyperparameters values of a ML algorithm and the size of the dataset portion to consider. Also in (Candelieri et al., 2020) an HPO task is considered, with two information sources, related to a large dataset and a small portion of it, respecitevely. All these approaches resulted more cost-efficient than using BO for HPO performed on the large dataset only, without significant degration of the final ML model's accuracy.

\subsection{Cost-Aware Optimization}
Although the quoted MISO and multi-fidelity approaches generalize from \textit{sample efficiency} to \textit{cost efficiency} -- thus, they are ``aware of costs'' -- they assume that the query cost of any source is constant over the search space. However, this is not true in practice, even using a single source, as empirically demonstrated in (Lee et al., 2020) according to the distribution of the time required to evaluate a set of 5000 randomly selected hyperparameter configurations, for five common ML algorithms. This is also confirmed in this paper (Sect. \ref{sec:task}), relatively to 1000 hyperparameters configurations for two ML algorithms on three datasets.

The seminal work aimed at making BO \textit{cost-aware} is (Snoek et al., 2012), which proposes to penalize the acquisition function, specifically Expected Improvement (EI), by the location-dependent cost, $c(x)$, leading to the \emph{EI-per-unit of cost}: $EI_{pu}(x)=EI(x)/c(x)$. However, $EI_{pu}(x)$ is basically driven by the query cost, biased towards cheap locations and, therefore, it performs well only when optima are relatively cheap. To overcome this undesired behviour, (Lee et al., 2020) proposes CArBO (Cost Apportioned BO), consisting of two consecutive stages: ($i$) a cost-effective selection of initial locations to query  (i.e., cost-effective initial design) and ($ii$) a \emph{cost-cooling} strategy where the penalty associated to the cost in $EI_{pu}(x)$ is modulated according to the query cost incurred so far. More precisely, $EI$-$cool(x)=\frac{EI(x)}{c(x)^\alpha}$, with $\alpha$ the \emph{cooling factor} defined as $\alpha=(\tau-\tau_n)/(\tau-\tau_{init})$, with $\tau$ the overall ``budget'' (i.e., maximum cumulated query cost), $\tau_n$ the cost cumulated up to the current iteration $n$, and $\tau_{init}$ the cost of the initial design. Finally, in CArBO, $c(x)$ is modelled through a warped GP.

Another recent paper (Paria et al., 2020) has proposed an approach whose cost-aware acquisition function is based on Information Directed Sampling (IDS) (Russo et al., 2014), a principled mechanism to balance \emph{regret} and \emph{information gain}. The proposed cost-aware acquisition function, namely CostIDS, also balances \emph{cost}, along with \emph{regret} and \emph{information gain}. However, an additional constraint is introduced, in optimizing CostIDS, to avoid that extremely cheap points are chosen repeatedly without any significant increase in information.

\subsection{Our contributions}
To the authors’ knowledge, this is the first paper addressing a multi-information source setting in which both fidelity and query cost of the sources are \emph{black-box} and \emph{location-dependent}.

Thus, we named our approach \textbf{MISO-wiLDCosts}: \textbf{MISO} \textbf{wi}th \textbf{L}ocation-\textbf{D}ependent \textbf{Costs}, whose goal is to solve (\ref{eq:globopt}), while keeping the cumulated query cost as small as possible, given a set of $S$ information sources, $\{f_s(x)\}_{s=1:S}$, whose approximation quality and query costs, $c_s(x)$, are black-box and location dependent.

We provide an empirical evaluation of MISO-wiLDCosts on an AutoML task: HPO of two ML classifiers on three datasets of different sizes.

\section{Gaussian Process regression}
\label{sec:gp}
GP modelling (Williams and Rasmussen, 2006) ---- INSERIRE GRAMACY---- is a non-parametric kernel-based learning method for probabilistic regression and classification. A GP regression model is a random function $f:\Omega \rightarrow \Re$ with output drawn from a multivariate normal distribution, formally $f(x) \sim \mathcal{N}(\mu(x),\sigma^2(x))$. In BO, GP regression is usually adopted as probabilistic rurrogate model, fitted on the $n$ function evaluations performed so far. Let $\mathbf{X_{1:n}}=\{x^{(1)},...,x^{(n)}\}$ and $\mathbf{y}=\{y^{(1)},...,y^{(n)}\}$ denote, respectively, the $n$ locations and the associated observed values, then fitting the GP means to compute the posterior mean and variance of the multivariate normal distribution as follows:
\begin{align}
\begin{split}
\label{eq:gpfit}
\mu(x)&=\textrm{k}(x,\mathbf{X_{1:n}}) \big[\mathbf{K}+\lambda^2\mathbf{I}\big]^{-1}\mathbf{y}\\
\sigma^2(x)&=k(x,x)-\textrm{k}(x,\mathbf{X_{1:n}}) \big[\mathbf{K}+\lambda^2\mathbf{I}\big]^{-1}\textrm{k}(\mathbf{X_{1:n}},x)
\end{split}
\end{align}
where $\lambda^2$ is the variance of a zero-mean Gaussian noise in the case of noisy observations (i.e., $y^{(i)}=f(x^{(i)})+\varepsilon$, with $\varepsilon \sim \mathcal{N}(0,\lambda^2)$), $\mathbf{K} \in \Re^{n \times n}$, such that $\mathbf{K_{ij}}=k(x_i,x_j)$, with $k$ a \emph{kernel} function modelling the covariance in the GP. Finally, $\textrm{k}(x,\mathbf{X_{1:n}})$ is vector whose $i$-th component is given by $k(x,x_i)$ (for completeness, $\textrm{k}(\mathbf{X_{1:n}},x)=\textrm{k}(x,\mathbf{X_{1:n}})^\top$).

The most widely adopted kernels are Squared Exponential (aka Gaussian), Mat\'ern, Power Exponential and Exponential (aka Laplacian). Each kernel implies a different prior on the structural properties of the sample paths of the latent function under the GP, such as differentiability. Moreover, each kernel has its own hyperparameters to adjust the GP's posterior depending on observations: GP's hyperparameters are usually tuned via Maximum Log-likelihood Estimation (MLE) or Maximum A Posteriori estimation (MAP).

In this paper, the Matern $3/2$ kernel is adopted:
\begin{equation}
k_{m_{3/2}}(x,x')=\sigma_k^2\Big(1+\frac{\sqrt{3}r}{\ell}\Big) e^{-\frac{\sqrt{3}r}{\ell}}
\end{equation}
with $r=||x-x'||$. The kernel's hyperparameters $\sigma_k^2$ and $\ell$, namely the kernel amplitude and the characteristic length scale, are estimated via MLE.

\section{MISO-wiLDCosts}
\subsection{Modelling sources and costs}
\label{sec:sources_and_costs}
Let $y_s$ denotes a function evaluation performed on the source $s$, where $y_s = f_s(x)+\varepsilon_s$ and  $\varepsilon_s$ is a zero-mean Gaussian noise associated to that source, formally $\varepsilon_s \sim \mathcal{N}(0, \lambda^2_s)$.
Let $z_s$ denotes the (black-box) query cost to observe $y_s$, that is $z_s = c_s(x)+\delta_s$, with $\delta \sim \mathcal{N}(0,\zeta^2_s)$.

It is important to remark that, by definition of query cost, $y_s$ and $z_s$ are not \emph{decoupled}, in the sense that $y_s$ cannot be observed without paying $z_s$, and vice-versa.

Let $D_s=\{x^{(i)},y_s^{(i)},z_s^{(i)}\}_{i=1,\dots,n_s}$ denotes the dataset collecting all the relevant information along the optimization process, with $n_s$ the number of function evaluations performed on the source $s$.
The following two projections of $D_s$ are considered:
\begin{equation}
F_s=\{(x^{(i)},y_s^{(i)})\}_{i=1,\dots,n_s}
\end{equation}
namely the \emph{function evaluations dataset}, storing locations queried and function values observed, and
\begin{equation}
C_s=\{(x^{(i)},z_s^{(i)})\}_{i=1,\dots,n_s}
\end{equation}
namely the \emph{query costs dataset}, storing locations queried and query costs. 

The two sets, $F_s$ and $C_s$, are used to fit two GPs, namely $\mathcal{F}_s(x)$ and $\mathcal{C}_s(x)$, according to (\ref{eq:gpfit}) and modelling:
\begin{equation}
f_s(x) \sim \mathcal{N}(\mu_s(x),\sigma_s^2(x) )
\end{equation}
and
\begin{equation}
c_s(x) \sim \mathcal{N}(p_s(x),q_s^2(x))
\end{equation}
with $s=1\dots,S$ and where $p_s(x)$ and $q_s(x)$ are again mean and standard deviation, but different symbols are used to distinguish them from $\mu_s(x)$ and $\sigma_s(x)$.

\subsection{Augmented Gaussian Process}
\label{sec:agp}
An Augmented Gaussian Process (AGP) (Candelieri et al., 2020) is aimed at generating a single model over all the $f_s(x)$ depending on the simplified \emph{model discrepancy} measure, which is devoted to measure the difference between two GPs approximating two different sources:
\begin{equation}
\label{eq:discrepancy}
\eta(\mathcal{F}_s,\mathcal{F}_{\bar s},x)=|\mu_s(x)-\mu_{\bar s}(x)|
\end{equation}
with $s,\bar s=1,\dots,S$ and $s \neq \bar s$.

In MISO-wiLDCosts, we assume that $f_1(x)$ is the preferred source, because, for instance, it coincides with $f(x)$ or it is supposed to provide the most accurate approximation for it. Fitting the AGP requires to \emph{augment} the set of function evaluations performed on the preferred source (i.e., the $F_1$ set), with function evaluations performed on other sources and whose function value is sufficiently close to the prediction provided by $\mathcal{F}_1(x)$. More formally, the set of augmenting locations, namely $\bar F$, is given by.
\begin{equation}
\bar F = \{(x, y_s ) \in F_s, s=2,\dots,S : \eta(\mathcal{F}_s,\mathcal{F}_1, x) < m \sigma_1(x) \}
\end{equation}
where $\eta(\mathcal{F}_s,\mathcal{F}_1,x)$ is computed according to (\ref{eq:discrepancy}) and $m$ is a techincal parameter, usually set to $m=1$, meaning that only evaluations falling into $\mu_1(x) \pm \sigma_1(x)$ are included in $\bar F$.

The final set of \emph{inducing locations} to use for fitting the AGP is given by $\widehat F = F_1 \cup \bar F$, and the resulting AGP is denoted by $\widehat \mathcal{F}(x)$, such that:
\begin{equation}
f(x) \sim \mathcal{N}(\widehat \mu_s(x), \widehat \sigma_s^2(x) )
\end{equation}
conditioned to $\widehat F$ according to (\ref{eq:gpfit}).

\subsection{Selecting next source-location to query}
To select the next source and locations to query, we started from the acquisition function proposed in (Candelieri et al., 2020) according to the AGP model. However, we have significantly modified it to consider the location dependent query cost of each source. The resulting acquisition function is based on the well-known Lower Confidence Bound (LCB), whose convergence proof under an appropriate scheduling of it's technical parameter, $\beta^t$, is given in (Srinivas et al., 2012).
The acquisition function proposed in this paper is defined as follows: 
\begin{equation}
\label{eq:acquisition}
(s', x') = \underset{ \substack{s=1,\dots,S\\x \in \Omega \subset \Re^d}}{\arg \max} \;\; \frac{\widehat y^+ -\Big[\widehat \mu(x) - \sqrt{\beta^t} \widehat \sigma(x)\Big]}{1+\overset{\triangle}{c_s}(x) \; \eta(\widehat \mathcal{F}, \mathcal{F}_s, x)}.
\end{equation}
where $\widehat y^+$ is the lowest function value of the inducing locations set $\widehat F$, that is $\widehat y^+ = \underset{(x,y) \in \widehat F}{\min}\{y\}$, and $\overset{\triangle}{c_s}(x)$ is an estimation of the query cost for the source $s$.

Therefore, the numerator of (\ref{eq:acquisition}) is the most optimistic improvement, with respect to $\widehat y^+$, depending on the AGP's LCB, while denominator consists of two \emph{source-and-location-dependent} penalization terms: $\overset{\triangle}{c_s}(x)$ and the model discrepancy between the AGP $\mathcal{\widehat{F}}(x)$ and the GP $\mathcal{F}_s(x)$ modelling the source $s$.

The location dependent query cost is modelled by the GP $\mathcal{C}_s(x)$ and a risk-averse attitude is considered, so $\overset{\triangle}{c_s}(x)$ is given by the upper confidence bound of $\mathcal{C}_s(x)$, that is the most pessimistic estimation of the cost to query $f_s(x)$. Formally $\overset{\triangle}{c_s}(x) = \max\{0,p_s(x) + q_s(x)\}$, to deal with negative values of the upper confidence bound, possibly due to the GP's approximation.


Finally, as reported in (Candelieri et al., 2020), there is the possibility that solving (\ref{eq:acquisition}) could lead to choose a pair $(s',x')$ whose location $x'$ is very close to a previously evaluated location on the source $s'$, leading to the ill-conditioning of the matrix $\big[\mathbf{K}+\lambda^2\mathbf{I}\big]$ in (\ref{eq:gpfit}) and, consequently, to the impossibility to update $\mu_s(x)$ and $\sigma^2_s(x)$. The correction propsed to avoid this undesired behaviour is:

\textbf{Correction}: \emph{If $\exists (x,y) \in F_{s'} : ||x - x'|| < \delta$ then set $s' \leftarrow 1$ and choose $x'$ as follows:}
\begin{equation}
\label{eq:correction}
x' \leftarrow \underset{x \in \Omega}{\arg \max} \; \sigma_1(x) 
\end{equation}
The idea is to choose an alternative $x'$ by ``investing'' the available budget on \emph{exploration} in order to improve the AGP at the next iteration by querying the most expensive source. Indeed, as stated in (Srinivas et al., 2012), selecting the location associated to the highest prediction uncertainty is a good strategy for \emph{function learning} (aka \emph{function approximation}), whose goal is to efficiently explore the search space to obtain an accurate approximation of $f(x)$ within a limited number of queries.

\section{Experimental Setting}
\subsection{Task description}
\label{sec:task}
To evaluate MISO-wiLDCosts, we considered a core application, that is the HPO of a ML algorithm. This task has been widely addressed via MISO/multi-fidelity optimization, such as in (Poloczek et al., 2017; Ghoreishi and Allaire, 2019; Candelieri et al., 2020; Swersky et al., 2013; Klein et al., 2017), as well as cost-aware optimization (Snoek et al., 2012; Paria et al., 2020; Lee et al., 2020).
Three binary classification datasets have been considered, different for size and number of features:
\begin{itemize}
\item SPLICE(3)\footnote{https://www.openml.org/d/1579} -- related to primate splice-junction gene sequences with associated imperfect domain theory, and used in (Lee et al., 2020). This dataset consists of 3175 instances, 60 (numeric) features plus the (binary) class label.
\item SVMGUIDE(1)\footnote{https://www.csie.ntu.edu.tw/~cjlin/libsvmtools/data\\sets/binary.html\#svmguide1} -- related to an astroparticle application from Jan Conrad of Uppsala University, Sweden (Chih-Wei et al., 2003). This dataset consists of 7089 instances, 4 (numeric) features plus the (binary) class label.
\item MAGIC GAMMA TELESCOPE\footnote{https://archive.ics.uci.edu/ml/datasets/magic+gam\\ma+telescope} -- generated through a Monte Carlo simulation software, namely Corsika, described in (Heck et al., 1998). The dataset has been used in (Candelieri et al., 2020) and consists of 19020 instances, 10 (numeric) features plus the (binary) class label.
\end{itemize}
All the features of the three datasets have been preliminary scaled in $[0,1]$.

The following ML classification algorithms, and associated hyperparameters, have been considered for the HPO task on the three datasets:
\begin{itemize}
\item Support Vector Machine classifier (C-SVC) with Radial Basis Function (RBF) kernel (Scholkopf and Smola, 2001). The hyperparameters to optimize are: the SVC's regularization term $C \in [10^{-2},10^2]$ and $\gamma \in [10^{-4}, 10^4]$ of the RBF kernel (i.e., $k(a,a')=e^-\frac{||a-a'||^2}{2\gamma^2}$, with $a \neq a'$ two instances within the dataset). 
\item Random Forest (RF) classifier (Goel and Abhilasha, 2017). The hyperparameters to optimize are the number of decision trees of the forest, $n_{trees} =\{300,\dots,700\}$, and the number of features subsampled to generate every tree, $m_{try} \in \big[ \lfloor 0.25 \times m_{feat}\rceil, \lfloor 0.75 \times m_{feat} \rceil \big] $, with $m_{feat}$ the number of features in the original dataset. The symbol $\lfloor . \rceil$ represents the rounding operation to the clostest integer value. 
\end{itemize}
The goal is to identify, for each dataset and for each ML classifier, the hyperparameters values minimizing the mis-classification error (mce), computed:
\begin{itemize}
\item on 10 fold cross validation (mce-FCV) for the C-SVC
\item and Out-Of-Bag (mce-OOB) for the RF classifier
\end{itemize}
while keeping the cumulated query cost as small as possible.

\begin{figure*}[h!]
\centering
\centerline{\fbox{\includegraphics[width=1.0\textwidth]{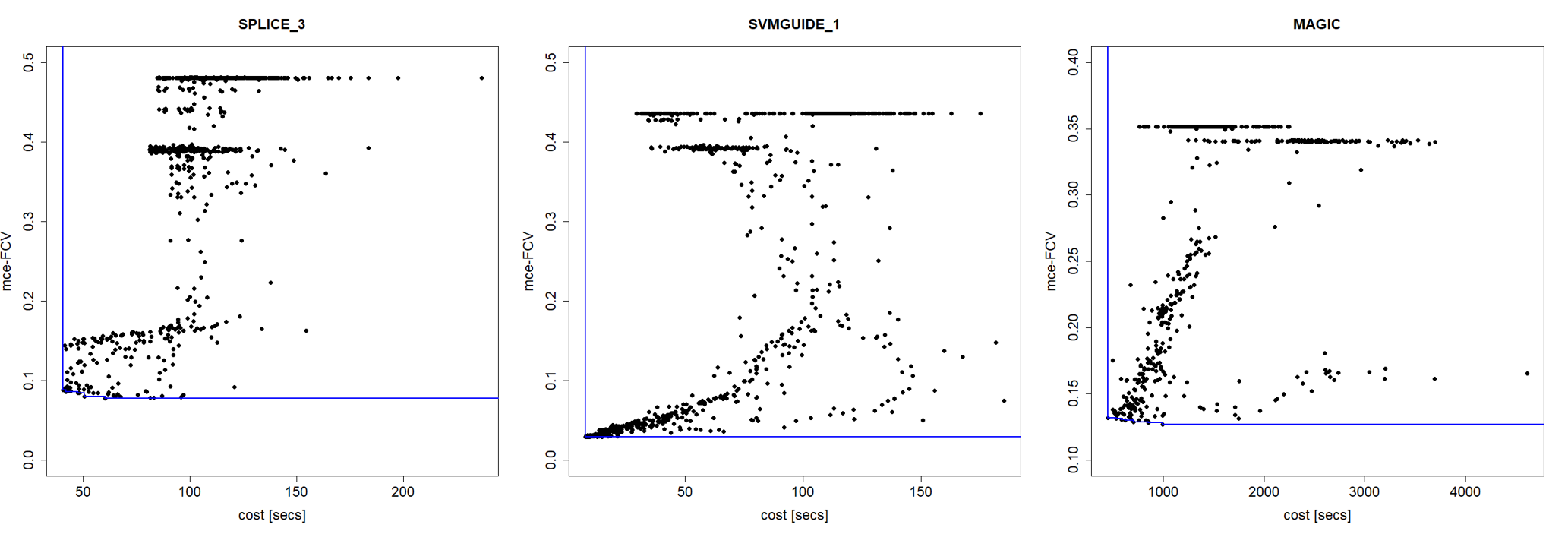}}}
\caption{Query cost and mce-FCV of 1000 randomly sampled C-SVC's hyperparameters configurations on three datasets: SPLICE(3) (left), SVMGUIDE(1) (middle) and MAGIC (right). The blue line can be assimilated to a Pareto frontier minimizing both mce and cost. In this case, the minimum mce should be also cheap.}
\label{fig:1}
\end{figure*}

\begin{figure*}[h!]
\vspace{.3in}
\centering
\centerline{\fbox{\includegraphics[width=1.0\textwidth]{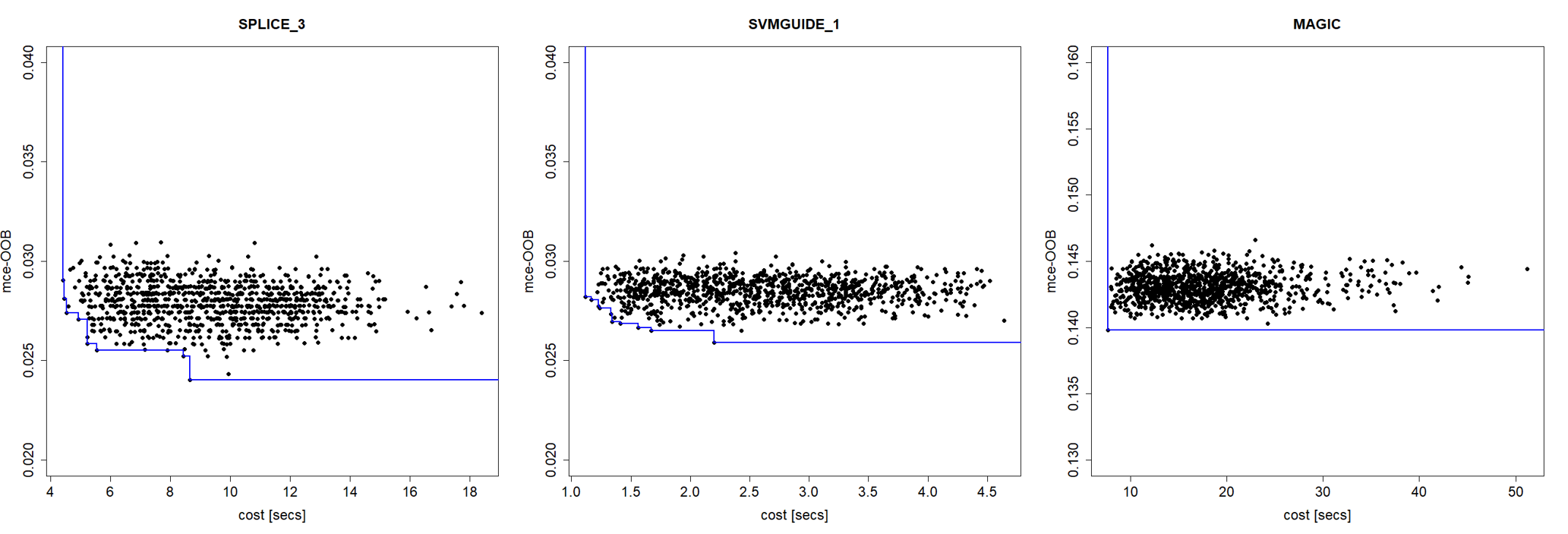}}}
\caption{Query cost and mce-OOB of 1000 randomly sampled RF's hyperparameters configurations on three datasets: SPLICE(3) (left), SVMGUIDE(1) (middle) and MAGIC (right). The blue line can be assimilated to a Pareto frontier minimizing both mce and cost. For the SPLICE(3) and SVMGUIDE(3) datasets, the mimimum mce is not associated to the cheapest RF's hyperparameters configurations.}
\label{fig:2}
\end{figure*}

Figure \ref{fig:1} and Figure \ref{fig:2} show, respectively for the C-SVC and the RF classifier, the query cost and the mis-classification error of 1000 hyperparameters configurations, randomly sampled via LHS, for each one of the three datasets. The blue line can be assimilated to a Pareto frontier, minimizing both mce and query cost, and it is used to make more evident the relation between the optimal mce and its cost.

First, when the two figures are compared, C-SVC's and RF's hyperparameters configurations have completly different mce values and query costs. This is mainly due to both the high computational cost for training an SVM classifier (i.e., number of instances powered to three) and differences in the computation of mce (i.e., on 10 FCV for C-SVC and OOB for RF). Then, C-SVC's mce values vary in a larger range than RF's ones. Therefore, the two classification algortihms can be considered two different representative cases.
With respect to C-SVC (Figure \ref{fig:1}), the sampled hyperparameters configurations show that (\textit{i}) the optimum mce should be ``cheap'' -- especially for the SVMGUIDE(1) dataset -- but (\textit{ii}) it could be difficult to reach, according to the small number of configurations around its minimum observed value, for all the three datasets. On the contrary, the minimum mce observed for the RF classifier is not associated to the cheapest hyperparameters configurations in the case of SPLICE(3) and SVMGUIDE(1) datasets (Figure \ref{fig:2}). Therefore, a lower mce can be achieved by more expensive RF classifiers, on these two datasets (even if it would be not significantly different from the average).

\subsection{Sources setup}
In MISO and multi-fidelity optimization for HPO, information sources are typically associated to small portions of the large original dataset. A similar idea is followed in this paper: each one of the three datasets was divided into 10 stratified subsets, then riaggregated to generate the following five sources:
\begin{itemize}
\item $f_1(x)=f(x)$, mce-FCV and mce-OOB on the entire dataset, respectively for C-SVC and RF;
\item $f_2(x)$, mce-FCV and mce-OOB on the 40\% of the original dataset (merging the first 4 subsets);
\item $f_3(x)$, mce-FCV and mce-OOB on the 30\% of the original dataset (merging subsets 5 to 7);
\item $f_4(x)$, mce-FCV and mce-OOB on the 20\% of the original dataset (merging subsets 8 and 9);
\item $f_5(x)$, mce-FCV and mce-OOB on the 10\% of the original dataset (just the subset 10)
\end{itemize}

According to this experimental setup, $f_2(x)$ to $f_5(x)$ rely on subsets that are not overlapping among them. Although this could imply a hierarchial organizations of the sources, we assume to do not anything about the composition and nature of the sources, they are completly black-box for the purposes of the study.
 
\subsection{MISO-wiLDCosts setup}
A set of 5 initial locations (i.e., hyperparameters values) sampled via Latin Hypercube Sampling (LHS), on each source, are used to initialize MISO-wiLDCosts. Ten independent runs are performed, for each dataset and for each classification algorithm, to mitigate the effect of random initialization.

At a generic iteration of MISO-wiLDCosts, all the GP models -- $\mathcal{F}_s(x)$, $\mathcal{C}_s(x)$ and $\widehat \mathcal{F}(x)$ -- are updated conditioned to the function values and costs observed so far. Then, the next $(s',x')$ to query is selected according to (\ref{eq:acquisition}), and in case (\ref{eq:correction}). Models updating and evaluation of the next selected source-location are iterated until fifty function evaluations are performed.

As final solution of each run, MISO-wiLDCosts returns the hyperparameters values, $x^+$, associated to the $\widehat y^+$ obtained at the end of the process, that is $x^+: (x^+, \widehat y^+) \in \widehat F$. This requires to compute, one last time, the set of the AGP's inducing locations, $\widehat F$. Finally, since $x^+$ could be a location queried on a source different from $f_1(x)$, a further evaluation is performed to replace $\widehat y^+$ with the $f_1(x^+)$. Although this situation has been considred in implementing MISO-wiLDCosts, it never occurred in the experiments performed.

\subsection{Baseline}
Since this is the first paper, at the authors' knowledge, addressing simultaneously MISO and cost-aware optimization, a comparison with other approaches was not so straightforward. A reasonable choice could be to compare MISO-wiLDCosts with a cost-aware BO approach performed on $f_1(x)$, only. In this case, the state of the art is represented by CArBO (Lee et al., 2020), and its \textit{cost-cooling} strategy has been adopted as a baseline. It is important to remark that CArBO also implements a procedure to obtain a cost-effective initial design, which could be also included in MISO-wilDCosts. In this paper, just the cost-cooling of CArBO has been considered, while the initial designs are those sampled via LHS in the MISO-wiLDCosts experiments.
Cost-cooling requires to define the overall budget in terms of maximum cumulated query cost. We defined it to cover approximately the fifty function evaluations performed by MISO-wiLDCosts, leading to:
\begin{itemize}
\item HPO of C-SVC: 1 hour for SPLICE(3) and SVMGUIDE(1), and 3 hours for MAGIC.
\item HPO of RF: 15 minutes for SPLICE(3), 10 minutes for SVMGUIDE(1), and 1 hour and a half for MAGIC.
\end{itemize}
Just for a fair comparison, also for CArBO cost-cooling the optimization process was stopped at fifty evaluations, even if some residal budget was available.

\subsection{Computational setting}
MISO-wiLDCosts is developed in R. Experiments were run on a Microsft Azure virtual machine, H8 (High Performance Computing family)  Standard with 8 vCPUs, 56 GB of memory, Ubuntu 16.04.6 LTS.

\section{Results}
The main results of our experiments are summarized in Table \ref{table:1}, where the suffixes ``wild'' and ``cooling'' are used to distinguish between MISO-wiLDCosts and CArBO cost-cooling, respectively.
Unsurprisingly, cumulated query cost is significantly lower for MISO-wiLDCosts, according to the Wilcoxon's non-paramatric test for paired samples ($p$-value$<$0.001). The most relevant outcomes related to mce are:
\begin{itemize}
\item with respect to C-SVC, mce-wild is higher than mce-cooling, for SPLICE(3) ($p$-value$<$0.0001) and SVMGUIDE(1) datasets ($p$-value=0.01);
\item mce-wild is equal to mce-cooling in the case of HPO of C-SVC on the MAGIC dataset ($p$-value$<$0.001). Although the difference between cost-wild and cost-cooling is less relevant than in the other cases, it is anyway statistically significant ($p$-value$<$0.0001);
\item as far as HPO of RF is concerned, mce-wild and mce-cooling are basically the same, but MISO-wiLDcosts used less than 40\% of the cumulated cost required by CArBO cost-cooling, less than 20\% for the SPLICE(3) and SVMGUIDE(1) datasets.
\end{itemize}

\begin{table*}[t]
\caption{HPO results on three classification datasets and two ML classification algorithms: mce and cumulated query cost for 10 runs of MISO-wiLDCosts and CArBO cost-cooling.}
\label{table:1}
\begin{center}
\begin{tabular}{llcccc}
\textbf{Classifier}  &  \textbf{Dataset}  &  \textbf{mce-wild}  &  \textbf{mce-cooling}  &  \textbf{cost-wild [secs]}  &  \textbf{cost-cooling [secs]}\\
\hline \\
C-SVC  &  SPLICE(3)  & $0.168\pm0.062$  & $0.080\pm0.002$  &  $744.607\pm91.760$  &  $3646.908\pm30.841$\\
C-SVC  &  SVMGUIDE(1)	&  $0.037\pm0.014$  &  $0.029\pm0.000$  &  $572.063\pm96.167$  &  $3263.303\pm385.737$\\
C-SVC  &  MAGIC  &  $0.172\pm0.000$  &  $0.172\pm0.000$ &  $11050.060\pm522.669$  &  $11575.700\pm699.917$\\
RF  &  SPLICE(3)  & $0.025\pm0.003$  & $0.025\pm0.000$  &  $78.624\pm7.375$  &  $456.572\pm42.073$\\
RF  &  SVMGUIDE(1)	&  $0.026\pm0.003$  &  $0.027\pm0.000$  &  $25.230\pm1.490$  &  $133.070\pm8.460$\\
RF  &  MAGIC  &  $0.141\pm0.000$  &  $0.141\pm0.000$ &  $492.327\pm39.704$  &  $1360.064\pm349.545$\\
\end{tabular}
\end{center}
\end{table*}

These differences are more evident in Table \ref{table:2}, where ``delta mce'' is the difference between mce-wild and mce-cooling, and ``\%cost'' is given by $100 \times \frac{\text{cost-wild}}{\text{cost-cooling}}$. However, it is important to remark that our results cannot be considered a comparative analysis with CArBO, because our implementation does not include the cost-effective initial design proposed in CArBO.

\begin{table*}[t]
\caption{Comparison between MISO-wiLDCosts and CArBO cost-cooling.}
\label{table:2}
\begin{center}
\begin{tabular}{llcc}
\textbf{Classifier}  &  \textbf{Dataset}  &  \textbf{delta mce}  &  \textbf{\%cost} \\
\hline \\
C-SVC  &  SPLICE(3)  & $0.088\pm0.063$  & $20.43\%\pm2.60\%$\\
C-SVC  &  SVMGUIDE(1)	&  $0.008\pm0.014$  &  $17.68\%\pm2.99\%$\\
C-SVC  &  MAGIC  &  $0.000\pm0.000$  &  $95.64\%\pm5.04\%$\\
RF  &  SPLICE(3)  & $0.000\pm0.003$  & $17.26\%\pm1.31\%$\\
RF  &  SVMGUIDE(1)	&  $-0.001\pm0.003$  &  $19.04\%\pm1.78\%$\\
RF  &  MAGIC  &  $0.001\pm0.000$  &  $38.65\%\pm11.12\%$\\
\end{tabular}
\end{center}
\end{table*}

\begin{figure*}[h!]
\centering
\centerline{\fbox{\includegraphics[width=1.0\textwidth]{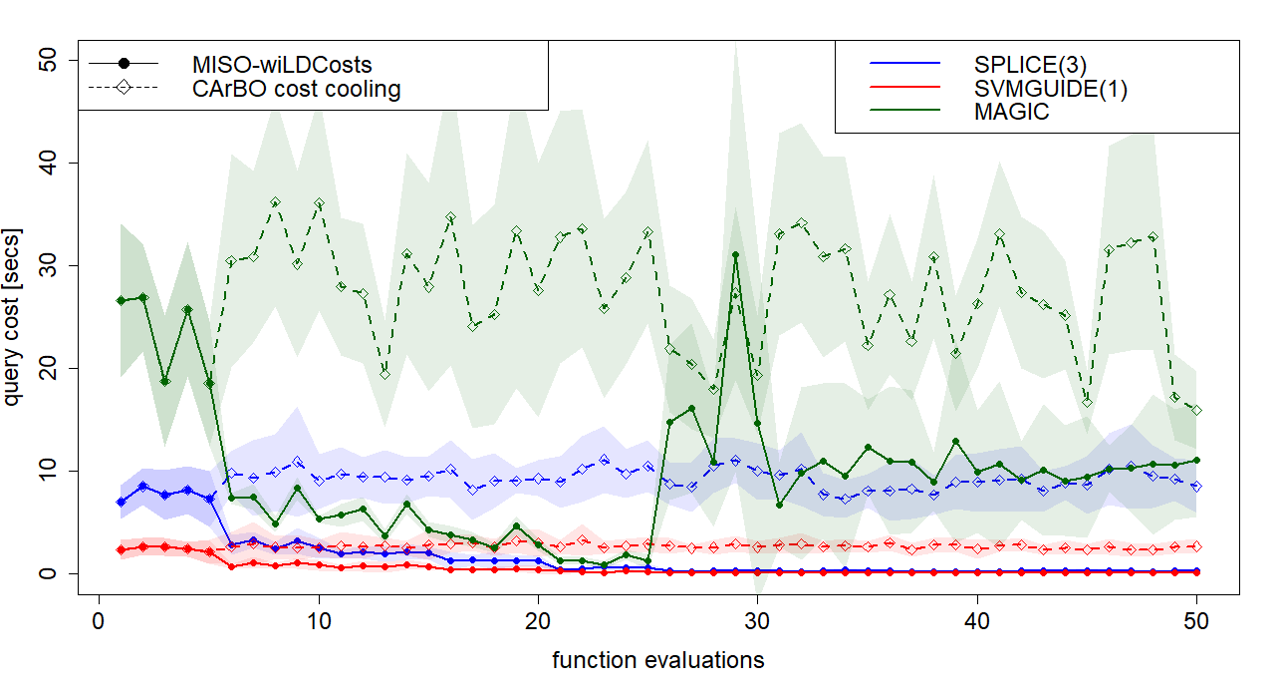}}}
\caption{RF classifier HPO on three datasets: query cost, by function evaluation, for MISO-wiLDCosts and CArBO cost-cooling. Lines and shaded area are, respectively, mean and standard deviation on 10 runs. Initial deisgn for CArBO cost-cooling consists of five hyperparameters configurations sampled via LHS, while for MISO-wiLDCosts it consists of five evaluations: five hyperparameters configurations on each one of the five sources.}
\label{fig:3}
\end{figure*}

Finally, just for illustrative purposes, Figure \ref{fig:3} shows the query cost incurred at each iteration of MISO-wiLDCosts and CArBO cost-cooling, separately (solid vs dashed lines): lines and shaded area are, respectively, mean and standard deviation computed on the 10 indepedent runs. For reasons of space and ease of viewing, the figure refers to HPO of RF, but the behaviour is analogous for HPO of C-SVC.

It is important to remark that just five initial hyperparameters configurations are used to initialize CArBO cost-cooling, while five for each source (i.e., twentyfive) are used to initialize MISO-wiLDCosts.

After the initial design, MISO-wiLDCosts mainly used ``cheap'' sources in the case of SPLICE(3) and SVMGUIDE(1) datasets, while it used all the sources, including $f_1(x)$, in the case of MAGIC dataset. This means that, in the case of SPLICE(3) and SVMGUIDE(1): ($i$) sources that were \emph{discrepant} from $f_1(x)$ had contributed in providing inducing locations of the AGP and/or ($ii$) any early convergence towards a specific source-location pair, $(s',x')$, did not occurred, thus correction (\ref{eq:correction}) was not (frequently) used. 

\section{Conclusions}
This paper shows that unifying MISO and cost-aware BO in a single framework can be accomplished obtaining good numerical performance.
MISO-wiLDCosts is the first MISO approach that is also aware of location-dependent costs within each source.
Its practical value is proven on a core application for both the MISO and cost-aware BO literature: the HPO of a ML algorithm on a large datset.
The computational results show that, compared to (single-source) cost-aware BO, MISO-wiLDCosts yields the same classification error at a significantly lower cumulated query costs.

\subsubsection*{Acknowledgements}
We greatly acknowledge the DEMS Data Science Lab, Department of Economics Management and Statistics (DEMS), University of Milano-Bicocca, for supporting this work by providing computational resources.


\end{document}